\documentclass[11pt]{article}

\usepackage[preprint]{acl}

\usepackage{times}
\usepackage{latexsym}
\usepackage{amsmath}

\usepackage[T1]{fontenc}
\usepackage[utf8]{inputenc}
\usepackage{multicol}

\usepackage{microtype}

\usepackage{inconsolata}

\usepackage{booktabs}
\usepackage{graphicx}
\usepackage{enumitem}
\usepackage{multirow}
\usepackage{float}
\usepackage{arydshln}
\usepackage[most]{tcolorbox}
\title{Uncertainty-Guided Latent Diagnostic Trajectory Learning for Sequential Clinical Diagnosis}

\author{
Xuyang Shen$^{1}$ \quad
Haoran Liu$^{2}$ \quad
Dongjin Song$^{1}$ \quad
Martin Renqiang Min$^{3}$ \\
$^{1}$University of Connecticut \\
$^{2}$Texas A\&M University \\
$^{3}$NEC Laboratories America
}

\begin{document}

\maketitle

\begin{abstract}
Clinical diagnosis requires sequentially acquiring and interpreting evidence under uncertainty. However, most Large Language Model (LLM) based diagnostic systems assume fully observed patient information, overlooking the sequential nature of diagnostic evidence acquisition. Even when diagnosis is formulated as a sequential decision process, learning effective diagnostic trajectories remains challenging because clinical datasets rarely provide explicit supervision indicating which sequences of diagnostic actions are most informative for reaching the correct diagnosis. To address this challenge, we propose Latent Diagnostic Trajectory Learning (LDTL), a framework that couples a planning LLM agent with a diagnostic LLM agent to learn sequential diagnostic strategies without explicit trajectory supervision. Specifically, LDTL treats diagnostic action sequences as latent trajectories and introduces a posterior distribution that favors trajectories yielding more informative diagnostic evidence. The planning agent is then trained to align with this posterior, encouraging coherent diagnostic action sequences that progressively reduce uncertainty. Experiments on the MIMIC-CDM benchmark demonstrate that LDTL improves diagnostic accuracy while requiring fewer diagnostic tests than comparable baselines. Further ablation studies highlight the importance of trajectory-level posterior alignment in achieving these improvements.

\end{abstract}
\section{Introduction}

\begin{figure}[t]
    \centering
    \includegraphics[width=0.48\textwidth,height=0.4\textheight,keepaspectratio]{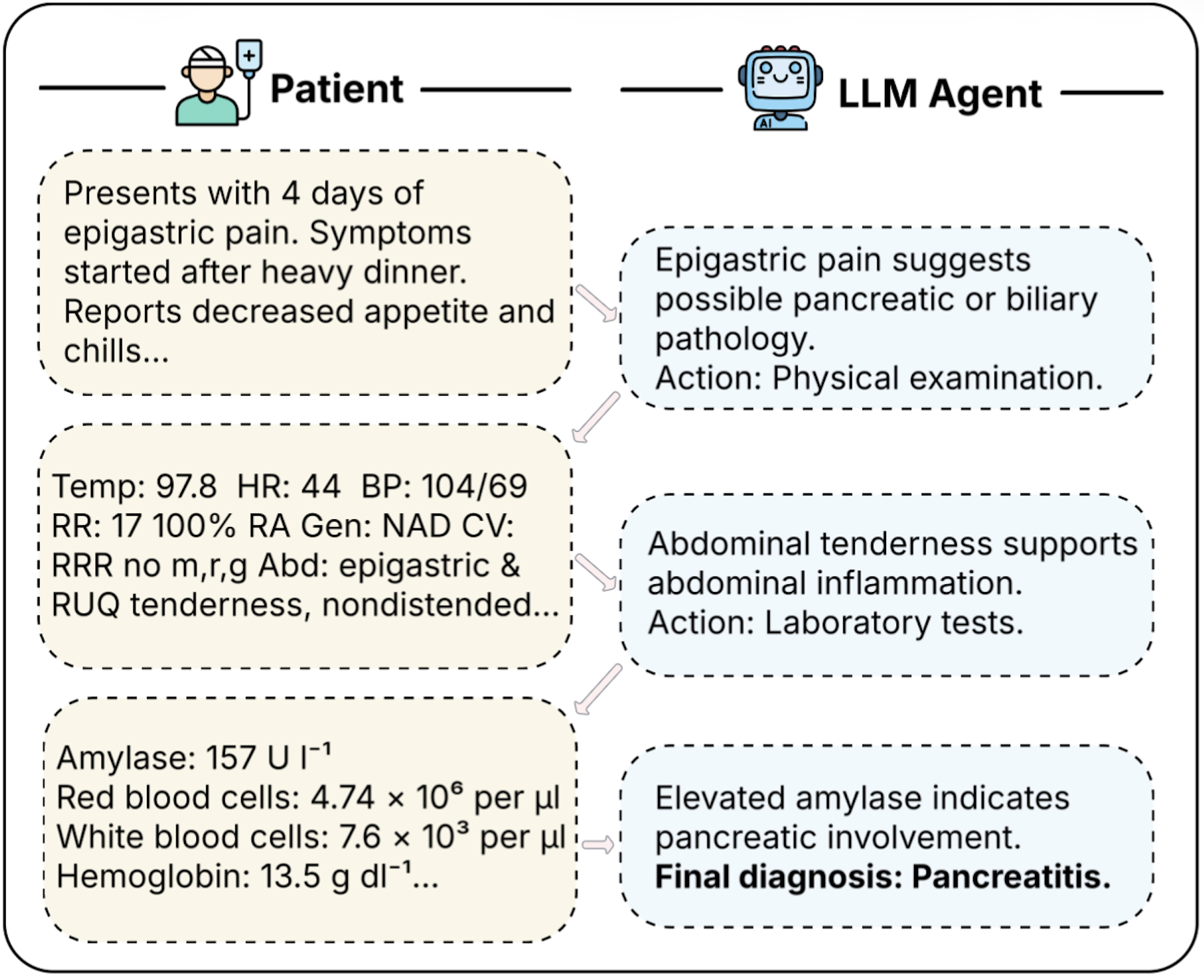}
    \caption{A real-world example of the  sequential clinical diagnosis system.}
    \label{fig:example}\vspace{-7mm}
\end{figure}

Accurate and timely diagnosis is a fundamental, sequential process in clinical practice. In real-world clinical settings, when a patient is admitted to a hospital, clinicians rarely have access to all necessary patient information at once. Instead, they typically start with partially observed information such as symptoms and medical history, and sequentially decide which examinations, laboratory tests, or imaging studies to order in the next step. Each newly acquired result helps refine the clinicians' hypothesis and progressively reduces uncertainty until sufficient confidence is reached for a final diagnosis decision. A concrete real-world example of the sequential clinical diagnosis system is shown in Figure \ref{fig:example}. Throughout this process, physicians must balance diagnostic accuracy with constraints of time, cost, and patient burden, making test selection a key part of clinical decision-making  \citep{rajpurkar2022ai}.

Recent advances in LLMs have demonstrated their strong capabilities in medical reasoning and diagnosis \citep{mcduff2025towards,brown2020gpt3,achiam2023gpt4,singhal2023medpalm2}. However, most existing approaches assume a full information setting in which all structured and unstructured patient data are available and accessible beforehand \citep{singhal2023medpalm, gilson2023usmle, achiam2023gpt4}. This assumption, however, is impractical due to the aforementioned real-world sequential clinical diagnosis process in which clinicians need to determine which tests to be ordered before making a confident diagnosis. Recent benchmarks and interactive frameworks,
such as MIMIC-CDM \citep{hager2024mimiccdm} and MediQ \citep{li2024mediq}, have begun to address this limitation by modeling sequential evidence acquisition under incomplete information. These settings highlight the importance of learning effective diagnostic action policies that adapt
to evolving patient information.

A natural approach is to formulate sequential diagnosis as an iterative agentic decision-making process, where an LLM agent selects diagnostic
actions and updates its predictions based on newly acquired evidence.
This perspective aligns with general LLM agent frameworks for multi-step reasoning and action selection \citep{yao2022react, shinn2023reflexion, wei2022chain, schick2023toolformer,yao2023tot, wang2024agents}. Recent studies have further explored reinforcement learning for cost sensitive test selection \citep{yu2023smddpo, sun2024edcopilot, schulman2017ppo} and hypothesis-driven clinical agents \citep{baniharouni2025lacdm, gao2026salus}. Other approaches, such as MedAction \citep{hsu2026medaction}, construct and filter multi-turn diagnostic trajectories to provide supervision for sequential diagnosis. Despite these advances, learning effective diagnostic trajectories remains challenging, as clinical records typically provide final diagnoses and test results without explicit preferred action sequences or desirable trajectories.

To address this challenge, we propose Latent Diagnostic Trajectory Learning (LDTL), a framework that learns diagnostic action preferences without requiring annotated optimal trajectories. LDTL consists of a planning LLM agent that selects diagnostic actions and a diagnostic LLM agent that evaluates accumulated evidence and determines when to stop. During training, the diagnostic agent estimates the improvement in the likelihood of the ground-truth diagnosis for candidate actions, which is used to construct a soft preference distribution for guiding the planner through KL-divergence alignment. During inference, the planner selects actions based solely on observed patient information. Our main contributions are summarized as follows:

\begin{itemize}[leftmargin=*, itemsep=1pt, topsep=0pt]

    \item We formulate sequential clinical diagnosis as a Latent Diagnostic Trajectory Learning (LDTL) problem and develop a framework integrating planning and diagnostic LLM agents for adaptive evidence acquisition and
    confidence-based diagnostic termination.

    \item We introduce an action preference learning strategy that leverages diagnostic improvement signals to construct soft supervision for the planning agent. This enables effective diagnostic action learning without requiring explicitly annotated optimal trajectories.

    \item Experiments on the MIMIC-CDM benchmark \citep{hager2024mimiccdm} demonstrate that LDTL improves diagnostic accuracy over comparable sequential baselines while reducing the number of evidence acquisition steps.

\end{itemize}

\section{Related Work}

Recent studies have explored sequential clinical diagnosis beyond traditional full-information settings. MIMIC-CDM \citep{hager2024mimiccdm} introduces a benchmark for sequential clinical decision-making, while MediQ \citep{li2024mediq} evaluates interactive question-asking under incomplete information. Other approaches investigate proactive multimodal consultation and multi-agent clinical decision-making \citep{liu2024medpmc,chen2025map}. More recently, MedAction \citep{hsu2026medaction} constructs and filters multi-turn diagnostic trajectories through simulated clinical interactions for supervised training. However, obtaining reliable supervision for intermediate diagnostic actions remains challenging, particularly when annotated desirable trajectories are unavailable.

Diagnostic test selection has also been studied as a cost-sensitive sequential decision problem. Reinforcement learning methods optimize diagnostic accuracy and resource efficiency \citep{sun2024edcopilot, yu2023smddpo, sun2026judging}. More recently, LA-CDM \citep{baniharouni2025lacdm} combines supervised and reinforcement learning to improve hypothesis generation, uncertainty estimation, and diagnostic efficiency, while Salus \citep{gao2026salus} employs multi-agent reinforcement learning with structured rewards for strategic diagnostic testing. Despite their effectiveness, these approaches require carefully designed optimization objectives and reward signals to balance diagnostic accuracy and evidence acquisition efficiency.

More broadly, LLM agent frameworks such as ReAct \citep{yao2022react} integrate reasoning with sequential action selection. Related studies in active learning and information-based data selection \citep{mackay1992information,settles2010active} investigate informative evidence acquisition, while confidence calibration methods \citep{guo2017calibration} address the reliability of model uncertainty estimates. Building on these directions, our method leverages diagnostic improvement signals to construct state conditioned action preferences, providing a tractable training signal for sequential diagnostic planning without explicit optimization over complete trajectories.

\section{Methodology}

\subsection{Problem Statement}

We formulate this clinical diagnosis as a sequential decision-making problem under partially observed information. For each patient, the initial observation is an incomplete clinical history \(h_0\), including symptoms and background information. At each step, the model may select a diagnostic action \(a_t \in \mathcal{A}\), where \(\mathcal{A}\) denotes the set of available tests or examinations. Executing \(a_t\) reveals new clinical evidence, which is incorporated into the patient state \(h_{t+1}\), consisting of the initial history together with all acquired test results until $t$ diagnostic steps. The process terminates at a patient-specific stopping time \(T\), when the model decides that sufficient evidence has been achieved for a final diagnosis.

A complete diagnostic process therefore induces a trajectory $z=(a_1,\dots,a_T)$,
and yields a disease prediction \(\hat y\) for the ground-truth label \(y\). The goal is to learn a policy that achieves accurate diagnosis while minimizing unnecessary test acquisition. This requires balancing predictive performance and diagnostic efficiency under uncertainty, without access to optimal diagnostic trajectories in training.

Unlike static diagnosis settings, where the full patient profile is observed upfront, sequential diagnosis requires the model to actively determine what information to acquire and when to stop. Accordingly, our formulation emphasizes learning over distributions of diagnostic trajectories rather than optimizing final prediction accuracy alone.

\begin{figure*}[t]
    \centering
    \includegraphics[width=0.95\textwidth]{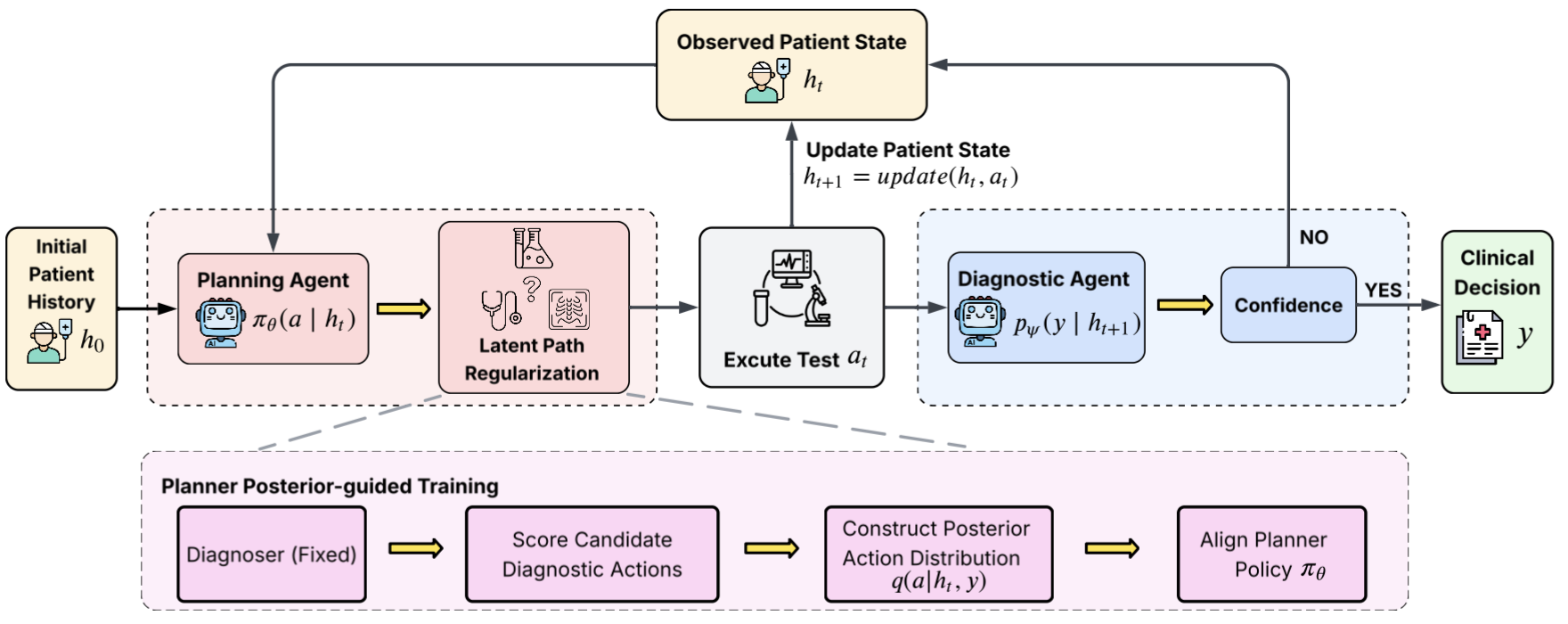}
    \caption{The proposed LDTL framework within a sequential diagnosis system. A planning LLM agent sequentially selects diagnostic tests and a diagnostic LLM agent updates the disease prediction and decides whether to stop based on the accumulated patient and test information.}
    \label{fig:framework}\vspace{-5mm}
\end{figure*}

\subsection{LDTL Framework}

Figure~\ref{fig:framework} illustrates the proposed LDTL framework, which consists of two collaborative LLM agents: a planning LLM agent and a diagnostic LLM agent. The planning LLM agent selects the next diagnostic action based on the current patient state, while the diagnostic LLM agent evaluates the updated evidence to produce a disease prediction and a termination decision.

At step \(t\), given the current state \(h_t\), the planning LLM agent chooses an action \(a_t \in \mathcal{A}\). After the corresponding test result is observed, the patient state is updated to \(h_{t+1}\). Conditioned on \(h_{t+1}\), the diagnostic LLM agent predicts a disease label and determines whether the accumulated evidence is sufficient to stop. If not, the updated state is passed back to the planning LLM agent to produce the next round of action.

Through this sequential process, the two LLM agents generate a patient-specific diagnostic trajectory \(z=(a_1,\dots,a_T)\). In the next subsection, we will introduce our learning objective for explicitly modeling the distribution over such trajectories.

\subsection{Diagnostic Trajectory Modeling}

The diagnostic process results in a sequence of information-acquisition actions determined by the planner. We treat this action sequence as a latent diagnostic trajectory that connects the initial patient state to the final diagnostic outcome \citep{chung2015vrnn}. Let \(h_0\) denote the initial patient history and \(y\) the ground truth disease label. 
Executing a trajectory sequentially reveals diagnostic evidence and updates the patient state according to $h_{t+1} = \text{Update}(h_t, a_t)$,
where the update operator incorporates the newly observed test result into the accumulated patient context.

The planner models a distribution over possible diagnostic trajectories through its conditional action policy:
\begin{equation}
p_\theta(z \mid h_0)
=
\prod_{t=1}^{T} \pi_\theta(a_t \mid h_t),
\label{eq:trajectory}
\end{equation}
where \(\pi_\theta(a_t \mid h_t)\) denotes the probability of selecting action \(a_t\) given the current patient state \(h_t\).

Given a trajectory \(z\), the diagnoser evaluates the accumulated evidence and produces a disease prediction. 
This interaction defines a diagnostic likelihood $p_\psi(y \mid h_0, z)$, which measures how well the collected diagnostic evidence supports the correct disease label.

Under this formulation, the diagnostic process can be viewed as marginalizing over a distribution of possible trajectories connecting the initial patient state to the final prediction. Different trajectories may provide varying diagnostic evidence, depending on the sequence of tests selected and the information they reveal. This perspective suggests that diagnostic trajectories should not be treated equally, and those providing stronger evidence for the correct disease should receive higher preference.

\subsection{Constructing the Trajectory Posterior}

Based on the above rationale, we introduce a trajectory-level preference distribution conditioned on the correct diagnosis to guide the planning LLM agent to pursue diagnostically informative action sequences. Intuitively, diagnostic trajectories that provide stronger evidence for the true disease should be assigned higher probability.

Formally, we define a trajectory preference distribution \(q(z \mid h_0, y)\) over candidate diagnostic trajectories as an energy-based model \citep{ziebart2008maxentirl}:
\begin{equation}\label{eq2}
q(z \mid h_0, y)
=
\frac{\exp(\beta S(z;h_0,y))}
{\sum_{z' \in \mathcal{Z}(h_0)} \exp(\beta S(z';h_0,y))},
\end{equation}

where $S(z)$ measures the diagnostic informativeness of trajectory $z$, reflecting how much the sequence of diagnostic actions improves confidence in the correct disease. In practice, assigning preference scores to complete diagnostic trajectories is nontrivial, as the
diagnostic value of each action depends on the evidence accumulated throughout the sequential process. Furthermore, clinical datasets do not provide explicit annotations of desirable diagnostic trajectories, making it difficult to directly supervise trajectory-level preferences.
\(\beta\) controls the sharpness of the distribution, and \(\mathcal{Z}(h_0)\) denotes the set of feasible trajectories starting from the initial patient state $h_0$. The score $S(z)$ reflects how informative the diagnostic trajectory is for identifying the correct disease. Trajectories that lead to more confident and accurate predictions are therefore assigned higher preference. This formulation provides a conceptual basis for assigning preferences to diagnostic trajectories without requiring annotated optimal paths.

\subsection{Action-Level Posterior}
Although trajectory level preferences characterize the diagnostic value of complete action sequences, learning such preferences remains challenging without explicit supervision of desirable diagnostic trajectories. Moreover, the informativeness of a diagnostic action depends on the current patient state and the evidence acquired through preceding actions. To address this challenge, we approximate trajectory level preferences through state conditioned action supervision, allowing the planning agent to learn informative diagnostic decisions without requiring annotated optimal trajectories.

Given the trajectory preference distribution defined in Eq.\ref{eq2}, the action level preference at step $t$ can be expressed by marginalizing over complete trajectories consistent with the current patient state:
\begin{equation}
q_{\mathrm{traj}}(a \mid h_t, y)
=
\frac{
\sum_{z \in \mathcal{Z}(h_t): z_t = a}
\exp(\beta S(z))
}{
\sum_{z' \in \mathcal{Z}(h_t)}
\exp(\beta S(z'))
},
\end{equation}
where $\mathcal{Z}(h_t)$ denotes the set of feasible trajectories consistent with the diagnostic history represented by $h_t$. This formulation captures the preference for an action based on its contribution to complete diagnostic trajectories.

However, evaluating this marginal requires assigning preference scores to complete trajectory continuations, for which explicit supervision is
unavailable. We therefore construct a tractable approximation using diagnostic improvement signals derived from the diagnostic LLM agent.

Given the current patient state $h_t$, the diagnostic LLM agent produces a predictive distribution over the candidate disease labels. Specifically, we obtain this distribution from the model output logits by applying a softmax over the scores corresponding to the four diseases, yielding $p_\psi(y \mid h_t)$.

For each candidate action $a$, we measure its diagnostic utility by the improvement in the log-probability assigned to the ground-truth diagnosis:
\begin{equation}
G_\psi(h_t,a;y)
=
\log p_\psi(y \mid h_{t+1}^{(a)})
-
\log p_\psi(y \mid h_t),
\end{equation}
where
$h_{t+1}^{(a)}=\mathrm{Update}(h_t,a)$
incorporates the observed result associated with action $a$. During training, the ground-truth diagnosis and recorded test outcomes are used to compute this improvement signal.

Actions that increase the model confidence in the correct diagnosis receive higher scores. We therefore define an approximate action level posterior over available candidate actions as
\begin{equation}\label{eq5}
\tilde{q}(a \mid h_t,y)
=
\frac{
\exp(G_\psi(h_t,a;y)/\tau)
}{
\sum_{a' \in \mathcal{A}(h_t)}
\exp(G_\psi(h_t,a';y)/\tau)
},
\end{equation}
where $\mathcal{A}(h_t)$ denotes the set of available candidate actions and $\tau$ is a temperature parameter controlling the smoothness of the distribution.

This approximation provides state-conditioned supervision without requiring explicit evaluation of complete diagnostic trajectories. By aligning the planner with these preferences across sequential states, the model is encouraged to select diagnostically informative actions. During inference, the planner directly selects actions based on the observed patient state,
without accessing the ground-truth diagnosis or unobserved test outcomes.

\subsection{Learning Strategy}
Joint training of the planning and diagnostic LLM agents introduces optimization challenges, as the planner’s supervision depends on a diagnostic model that evolves during training, leading to a moving target. To avoid this coupling and provide a stable reference for computing diagnostic improvement, we adopt a two-stage learning strategy.

\noindent \textbf{Stage 1}: Fine-tune diagnostic LLM agent.
We first fine-tune the diagnostic LLM agent on supervised patient records containing full diagnostic information. Given the complete set of diagnostic results, the model learns to predict the correct disease label and provide a confidence assessment for diagnostic termination.

This stage serves two purposes. First, it stabilizes the output format and probability distribution of the LLM for clinical diagnosis tasks. Second, it provides a reliable diagnostic reference model whose predictions can be evaluated under different information states. After convergence, the diagnostic LLM agent is frozen and used as a fixed module for planning LLM agent learning.

\noindent \textbf{Stage 2}: Posterior-guided planner LLM learning. By fixing the diagnostic LLM agent, we train the planning LLM agent to select diagnostic actions sequentially. As defined in Eq.~\ref{eq:trajectory}, the planner induces a trajectory distribution through its state-conditioned action policy $\pi_\theta(a \mid h_t)$.

At each step, we construct an approximate action level posterior $\tilde{q}(a \mid h_t,y)$ using diagnostic improvement signals from the frozen diagnostic LLM agent, as defined in Eq.~\ref{eq5}. This distribution provides supervision for the planner without requiring annotated optimal diagnostic trajectories.

The planning LLM agent is optimized by aligning its action policy with this posterior through
\begin{equation}
\mathcal{L}_{\mathrm{planner}}
=
\sum_{t=1}^{T}
D_{\mathrm{KL}}
\left(
\tilde{q}(\cdot \mid h_t,y)
\;\|\;
\pi_\theta(\cdot \mid h_t)
\right).
\end{equation}

This objective encourages the planner to assign higher probabilities to diagnostically informative actions across sequential patient states. Although optimization is performed at the action level, learning across successive states promotes effective evidence acquisition throughout the diagnostic process.

During training, the ground-truth diagnosis and recorded test outcomes are used to construct the action level posterior. During inference, the planning LLM agent selects actions directly according to its learned policy
$\pi_\theta(a \mid h_t)$, without access to the ground-truth diagnosis or unobserved test outcomes.

% The planning LLM induces a trajectory distribution through its conditional action policy as defined in Eq.~\ref{eq:trajectory}, where \(h_t\) denotes the patient state after incorporating previously observed test results. Using the action level posterior \(q(a_t \mid h_t, y)\) constructed in Eq.~\ref{eq5}, we convert trajectory level diagnostic preferences into step-wise supervision. The planner LLM is optimized by aligning its policy with this posterior through
% \begin{equation}
% \mathcal{L}_{\text{planner}} =
% \sum_{t=1}^{T}
% \mathrm{KL}\big(q(a \mid h_t, y) \;\|\; \pi_\theta(a \mid h_t)\big),
% \end{equation}
% which encourages the planner to align its action distribution with the posterior derived from diagnostic improvement. Although optimization is performed locally at each step, the posterior $q(a \mid h_t,y)$ aggregates information from trajectory level preferences, thereby imposing structural regularization over diagnostic paths.

% Through this two-stage training strategy, the diagnostic LLM agent provides a fixed probabilistic reference model, while the planning LLM agent learns to select informative diagnostic actions.  During training, the action posterior $q(a_t \mid h_t, y)$ is computed using the ground-truth diagnosis $y$ to provide supervision for the planning LLM agent.  During inference time, the planning LLM agent selects actions directly according to its learned policy $\pi_\theta(a_t \mid h_t)$ without access to the true label.

\section{Experiments}
In this section, we describe the experimental setup and evaluate LDTL through comparisons with baseline methods, ablation studies, efficiency analysis, and case studies.

\subsection{Experimental Setup}
\subsubsection{Dataset and Pre-processing}

We evaluate on MIMIC-CDM \citep{hager2024mimiccdm, goldberger2000physiobank}, a curated subset of MIMIC-IV \citep{johnson2023mimiciv} designed to support sequential clinical decision-making. The benchmark contains 2400 inpatient cases labeled with one of four abdominal conditions: appendicitis, cholecystitis, diverticulitis, pancreatitis. Each case contains multiple types of clinical data. The initial record includes patient history information (documented symptoms, comorbidities, and family history). In addition, diagnostic tests also provide physical examination notes, imaging reports covering CT, X-ray, ultrasound, and MRI modalities, as well as laboratory records spanning blood, urine, and microbiology tests. But not every diagnostic test is available for every patient. Details of the data preprocessing procedure and action-space construction are 
described in Appendix~\ref{appendix:data}.

\begin{table*}[t]
\centering
\small
\setlength{\tabcolsep}{4pt}
\caption{Performance comparison of our method LDTL and baseline methods. 
We report class-wise accuracies and F1-scores. 
ZS = zero-shot.}
\vspace{-2mm}
\begin{tabular}{lccccccccc}
\toprule
 & \multicolumn{5}{c}{Accuracy} 
 & \multirow{2}{*}{F1-Score} 
\\
\cmidrule(lr){2-6} 
Method 

& Appendicitis & Cholecystitis & Diverticulitis & Pancreatitis & Overall 
 \\
\specialrule{0.75pt}{0pt}{0pt}

SFT-all
& 97.9 & 93.1 & 90.4 & 90.7 & 94.2 
& 95.8 &
\\

\midrule

Fixed info (history) (ZS)
& 47.9 & 34.6 & 73.1 & 79.6 & 54.1 
& 58.8
\\

Fixed info (history + one test) (ZS)
& 50.5 & 47.7 & 76.9 & 72.2 & 58.3 
& 60.5 
\\

Fixed info (history + two tests) (ZS)
& 60.9 & 60.7 & 78.8 & 80.5 & 67.2
& 68.2 
\\

All info (ZS)
& 73.9 & 70.0 & 82.7 & \underline{87.9} & 76.9
& 76.7
\\

Planner random selection
& 82.8 & \underline{85.4} & \textbf{90.4} & 85.2 & \underline{84.8}
& 83.7 
\\

ReAct \citep{yao2022react}
& 90.2 & 79.7 & 66.7 & 62.9 & 74.9 
& 79.1 & 
\\

LA-CDM \cite{baniharouni2025lacdm}
& \underline{93.1 } & 83.6 & 75.0 & 73.5 & 81.3 
& \underline{84.1} 
\\

\midrule
\textbf{LDTL}
& \textbf{98.9} & \textbf{95.4} & \underline{88.8} & \textbf{87.9} & \textbf{93.4} 
& \textbf{91.7} 
\\

\bottomrule
\end{tabular}

\label{tab:main_results}\vspace{-4mm}
\end{table*}

\subsubsection{Evaluation Metrics}
We evaluate both diagnostic performance and decision efficiency. Diagnostic performance is measured using final prediction accuracy and F1 score at termination, where termination is defined as either make a diagnosis or reaching a predefined maximum number of decision steps. To assess efficiency, we report the number of evidence acquisition steps per case and the proportion of cases that terminate before reaching the maximum planning steps. These metrics allow us to examine the trade-off between diagnostic accuracy and information acquisition cost.

\subsubsection{Baseline Model}
We use Llama-3-8B \citep{dubey2024llama3} as the backbone model due to its strong general reasoning capability and open source availability. The 8B scale provides a practical balance between reasoning performance and computational efficiency. To the best of our knowledge, the model has not been pretrained on the MIMIC-CDM dataset. All experiments are conducted using publicly available pretrained weights without any additional pretraining on the dataset. Based on this backbone, we compare our method against several representative baselines that differ in how diagnostic information is acquired and utilized.

\noindent\textbf{One-shot full-information model.}
All available clinical evidence is provided to the base model at once, and a diagnosis is predicted without sequential decision making. This setting evaluates performance under complete information access.

\noindent\textbf{Static information strategies.}
Instead of adaptive test selection, we use fixed information subsets formed by combining the patient history with a predefined set of test categories. Specifically, we consider (i) history only, (ii) history plus exactly one test category (physical examination, laboratory tests, or imaging), and (iii) history plus two of the three test categories. The model receives these inputs in a single forward pass and produces a diagnosis. These baselines assess performance under non-adaptive but structured information acquisition.

\noindent\textbf{Random policy.}
At each decision step, a diagnostic action is sampled uniformly from the action space until termination. This baseline verifies the benefit of learned planning compared to uninformed exploration.

\noindent\textbf{Sequential agent baselines.}
We further compare with prior language agent decision frameworks, including the prompting based agent ReAct~\citep{yao2022react} and the reinforcement learning based method LA-CDM~\cite{baniharouni2025lacdm}.

\subsection{Main Results}

Table~\ref{tab:main_results} presents the diagnostic performance of LDTL compared with static and sequential baselines. We report class-wise accuracy, computed as the proportion of correctly predicted cases within each disease category, overall accuracy, and F1 score.

We first consider the SFT-all setting, where the diagnostic LLM agent is fine-tuned and evaluated with complete patient information. This full-information reference achieves 94.2\% overall accuracy and 95.8 F1 score, without requiring sequential test selection. In contrast, zero-shot static strategies show substantially lower performance under limited information. Using patient history alone yields 54.1\% accuracy, which increases to 58.3\% and 67.2\% when one or two test categories are included, respectively. Providing all available information without fine-tuning achieves 76.9\% accuracy, indicating that additional clinical evidence alone is insufficient to match the performance of task-specific fine-tuning.

\begin{table*}[htbp]
\centering
\small
\setlength{\tabcolsep}{5pt}
\vspace{3mm}
\caption{Ablation study on Llama3-8B. We compare random selection (RS), a state-conditioned planner without latent path regularization (w/o LP), and our full model with trajectory-level posterior alignment over five seeds.}\vspace{-2mm}
\begin{tabular}{lcccccc}
\toprule
Method & Acc & F1 & Appendicitis & Cholecystitis & Diverticulitis & Pancreatitis \\
\specialrule{1pt}{1pt}{1pt}

RS     & 85.4 $\pm$ 1.3 & 84.9 $\pm$ 1.3 & 83.5 $\pm$ 1.1 & 85.8 $\pm$ 1.5 & \textbf{90.9 $\pm$ 2.8} & 85.4 $\pm$ 2.1 \\
w/o LP & 87.9 $\pm$ 0.7 & 87.8 $\pm$ 0.8 & 94.0 $\pm$ 1.3 & 90.9 $\pm$ 1.4 & 81.1 $\pm$ 1.7 & 78.1 $\pm$ 0.9 \\
\midrule
LDTL   & \textbf{93.0 $\pm$ 0.4} & \textbf{92.4 $\pm$ 0.4} & \textbf{97.6 $\pm$ 0.5} & \textbf{95.2 $\pm$ 1.0} & 88.4 $\pm$ 2.5 & \textbf{86.0 $\pm$ 2.4} \\
\bottomrule
\end{tabular}

\label{tab:ablation}\vspace{-5mm}
\end{table*}

We next evaluate sequential decision-making frameworks. A random planner that selects diagnostic actions uniformly achieves 84.8\% accuracy. Unlike static strategies, sequential frameworks progressively incorporate newly acquired evidence into the patient state, allowing diagnostic predictions to be updated throughout the process. However, random action selection does not explicitly prioritize diagnostically informative evidence, potentially resulting in unnecessary tests or less effective diagnostic trajectories.

Among existing sequential agents, ReAct is a prompting-based approach, whereas LA-CDM employs task specific optimization to improve hypothesis generation and diagnostic action selection. LA-CDM achieves 81.3\% accuracy, outperforming ReAct at 74.9\%, suggesting the potential benefits of explicitly optimizing clinical decision making. Nevertheless, both methods remain below the random planner in overall accuracy on this benchmark. These results indicate that iterative reasoning and action selection alone do not necessarily ensure effective diagnostic evidence acquisition, highlighting the importance of learning informative action preferences.

Our proposed LDTL achieves 93.4\% overall accuracy and 91.7 F1 score, outperforming all evaluated sequential baselines and approaching the full information SFT-all reference. Specifically, LDTL improves accuracy by 8.6 and 12.1 percentage points over the random planner and LA-CDM, respectively. The improvements are particularly evident for appendicitis and cholecystitis, although performance gains are not uniform across all disease categories, with relatively lower accuracy on diverticulitis compared with random selection. Overall, these results suggest that aligning the planning policy with diagnostic outcome based action preferences facilitates more effective evidence acquisition and improves final diagnostic performance.

\begin{figure*}[htbp]
    \centering
    \includegraphics[width=\textwidth]{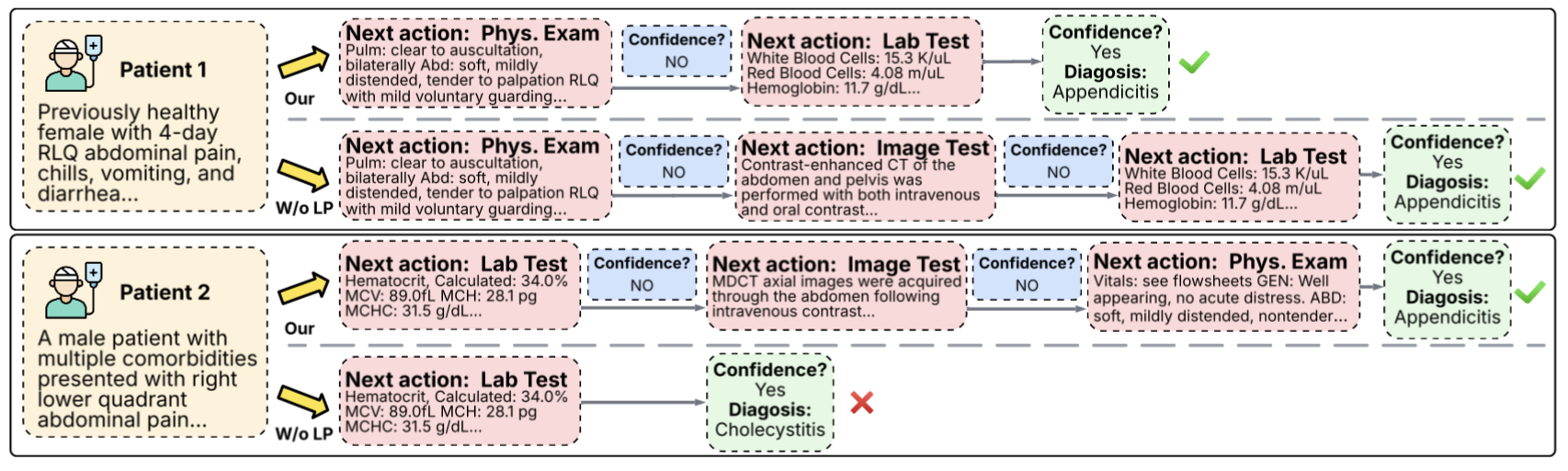}
    \caption{Case studies comparing our method LDTL with the variant without latent path regularization (w/o LP). }
    \label{fig:casestudy}\vspace{-5mm}
\end{figure*}

\subsection{Ablation Study}

To evaluate the effectiveness of posterior-guided planner learning, we compare three variants: (1) random selection (RS), which uniformly samples diagnostic actions; (2) a state-conditioned planner without posterior-guided supervision (w/o LP); and (3) our full LDTL framework. As shown in Table~\ref{tab:ablation}, results are reported as mean $\pm$ standard deviation over 5 random seeds. RS achieves 85.4$\pm$1.3 accuracy, while w/o LP improves to 87.9$\pm$0.7, demonstrating the benefit of learning state-conditioned diagnostic action selection.

Our full LDTL further improves accuracy to 93.0$\pm$0.4 and F1 to 92.4$\pm$0.4, outperforming w/o LP by 5.1 percentage points in accuracy. Unlike w/o LP, LDTL aligns the planning policy with diagnostic improvement-based action preferences, providing more informative supervision for sequential decision-making. These results demonstrate the effectiveness of posterior-guided learning in improving diagnostic performance.

\vspace{-3mm}
\subsection{Diagnostic Efficiency Analysis}
\vspace{-2mm}

\begin{figure}[t]
\centering
\includegraphics[width=0.97\linewidth]{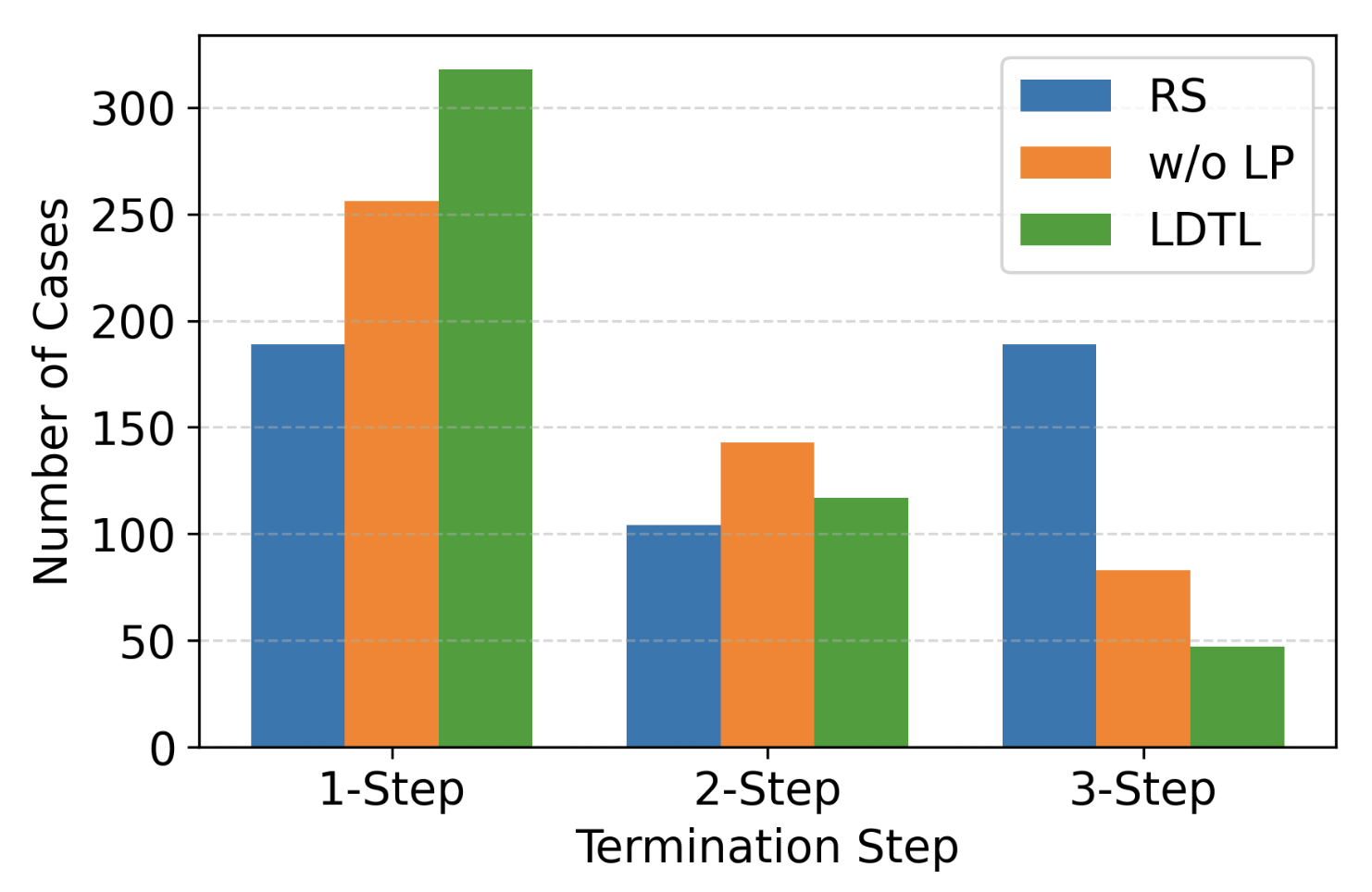}\vspace{-3mm}
\caption{Distribution of diagnostic termination steps on the test cases. Numbers indicate how many cases terminate after one, two, or three actions.}
\label{fig:termination_steps}\vspace{-5mm}
\end{figure}

Beyond final accuracy, we analyze how many diagnostic steps are required before termination across the test dataset. Figure~\ref{fig:termination_steps} reports the number of cases that terminate after one, two, or three actions for each method.

Our method achieves the highest proportion of one-step terminations while requiring the fewest three-step trajectories. This indicates that our model (LDTL) can more frequently identify decisive diagnostic evidence early and avoid unnecessary additional tests. In contrast, random selection exhibits a large number of three-step trajectories, reflecting inefficient action choices that fail to quickly reduce diagnostic uncertainty. The state conditioned planner without latent path regularization improves over random exploration but still requires more multi-step trajectories compared to our approach.

\begin{table}[t]
\centering
\small
\setlength{\tabcolsep}{5pt}
\vspace{3mm}
\caption{Step-conditioned accuracy and efficiency over five seeds. The table reports diagnostic accuracy conditioned on the termination step along with final accuracy.}\vspace{-2mm}
\begin{tabular}{lccc}
\toprule
Method & 1-step Acc & 2-step Acc & 3-step Acc  \\
\specialrule{1pt}{1pt}{1pt}

RS     & 85.2 $\pm$ 1.3 & 87.4 $\pm$ 4.0 & 83.4 $\pm$ 3.4  \\
w/o LP & 88.9 $\pm$ 1.0 & 90.9 $\pm$ 2.2 & 80.5 $\pm$ 2.8  \\
LDTL   & 93.2 $\pm$ 0.4 & 95.1 $\pm$ 2.3 & 84.7 $\pm$ 2.2 \\
\bottomrule
\end{tabular}

\label{tab:steps}\vspace{-2mm}
\end{table}

And Table \ref{tab:steps} further reports diagnostic accuracy conditioned on the termination step. LDTL achieves the highest accuracy at both 1-step and 2-step terminations, indicating the model is able to reach correct diagnoses with fewer steps. In contrast, random selection and the w/o LP variant exhibit lower accuracy at early steps, suggesting less reliable evidence acquisition. Overall, these observations suggest that posterior-guided action learning helps the planner prioritize informative diagnostic actions earlier, improving both diagnostic accuracy and evidence acquisition efficiency.

\vspace{-3mm}
\subsection{Case Study}
\vspace{-2mm}

To further understand the behavioral differences resulting from latent path regularization, we analyze representative diagnostic trajectories from the test set in Figure~\ref{fig:casestudy}.
\vspace{-2mm}

\paragraph{Efficient evidence acquisition.}
In Patient 1 with true label appendicitis, LDTL first performs a physical examination and then requests laboratory tests, after which the diagnostic LLM agent reaches sufficient confidence and terminates with the correct diagnosis. In contrast, the planning LLM agent without latent path regularization follows a less efficient trajectory, requesting an additional imaging test before obtaining the same laboratory evidence. This example illustrates that trajectory-level regularization helps prioritize diagnostically informative actions and avoids unnecessary tests.

\vspace{-3mm}
\paragraph{Avoiding premature termination.}
Patient 2 demonstrates a different failure mode. 
LDTL gradually collects evidence through laboratory tests, imaging, and physical examination before reaching a confident and correct diagnosis. Without latent path regularization, however, the planning LLM agent terminates after the first laboratory test and produces an incorrect prediction cholecystitis. This example shows that our method encourages more reliable evidence accumulation and prevents overconfident early decisions.

\vspace{-3mm}
\section{Conclusion}
\vspace{-2mm}
We propose Latent Diagnostic Trajectory Learning (LDTL), an uncertainty-guided framework for sequential clinical diagnosis that models diagnostic action sequences as latent trajectories. By leveraging diagnostic improvement signals from a diagnostic LLM agent, LDTL constructs a soft action preference distribution to guide the planning LLM agent without requiring annotated optimal diagnostic trajectories. Experiments on MIMIC-CDM demonstrate that LDTL improves diagnostic accuracy over comparable sequential baselines while reducing the number of evidence acquisition steps. These findings highlight the effectiveness of posterior-guided action learning for sequential clinical diagnosis under partial information.

\section*{Limitations}
Although our framework improves diagnostic decision-making in a simulated setting and obtains encouraging experimental results, it should not be used for real clinical decision support without extensive validation. Large language models may produce incorrect or misleading outputs, and clinical deployment requires strict human oversight. Our study has several limitations as well.

First, experiments are conducted on a single abdominal diagnosis benchmark with a limited number of disease classes and moderate dataset size. Although MIMIC-CDM simulates sequential evidence acquisition, it does not fully reflect the scale, diversity, and noise characteristics of real hospital environments. Evaluating the proposed sequential framework on broader clinical domains datasets remains an important direction for future work.

Besides, our current formulation defines the planner action space at the level of clinically meaningful test categories. This abstraction simplifies sequential decision modeling and ensures stable learning, but does not capture finer-grained decisions at the individual test level within each category. Extending the framework to more granular and heterogeneous action spaces may require additional mechanisms for structured action representation.

Finally, our framework assumes a fixed and predefined set of candidate diagnostic actions. In practice, clinical workflows may involve dynamically generated hypotheses, conditional test dependencies, and external knowledge retrieval. Incorporating open ended action generation and structured medical guidelines into the latent path formulation remains an important direction \cite{shafer2008conformal,vovk2005alrw}. In addition, the current framework uses a binary confidence signal for termination rather than a probabilistically calibrated estimate. Developing calibrated confidence measures for safer clinical decision-making remains an important direction for future research.

%\section*{Acknowledgments}


\newpage
\bibliography{main}

\clearpage
\appendix

\onecolumn
\nolinenumbers
\section{Prompts}
\label{prompts}

\subsection{Planning Agent Prompt}
% \begin{figure*}[t]
\begin{tcolorbox}[colback=gray!5,colframe=black,title=Planner Prompt]

\ttfamily\small\raggedright

You are an experienced emergency physician responsible for evaluating a patient who presents with abdominal symptoms. Your goal is to plan the diagnostic process by deciding what type of examination should be performed next.

At each step, you will receive the patient’s currently available clinical information, which may include symptoms, medical history, previously completed examinations, and their results. Based on this information, determine the most appropriate next diagnostic examination that would help clarify the patient's condition and reduce diagnostic uncertainty.

Your objective is to gather information that progressively narrows down the possible diagnoses and supports accurate clinical decision-making.

Possible diagnostic examinations include:

- laboratory tests  

- physical examination

- imaging studies  

When selecting the next step, consider what additional information would be most informative for distinguishing between possible diagnoses and guiding further clinical decisions. Avoid repeating examinations that have already been performed.

Examples:

Example 1

Patient information:  
A 45-year-old male presents with severe abdominal pain and fever.

Clinical consideration:  
Initial symptoms suggest an acute abdominal process, and physical examination is needed to further localize tenderness and assess severity.

Next diagnostic step:  
["physical"]

Example 2

Patient information:  
A 60-year-old male presents with chronic cough. Laboratory tests show normal CBC. Physical examination reveals wheezing.  
Completed examinations: ["lab", "physical"]

Clinical consideration:  
Since basic laboratory and physical findings are already available, imaging would provide additional evidence to further evaluate the underlying cause.

Next diagnostic step:  
["image"]

Now evaluate the following patient case.

Patient current information:  
\{input\}

Return the next diagnostic step using the following format:

["lab"], ["physical"], or ["image"]

\end{tcolorbox}
% \end{figure*}

\subsection{Diagnostic Agent Prompt}
% \begin{figure*}[t]
\begin{tcolorbox}[colback=gray!5, colframe=black,title=Diagnoser Prompt, breakable]

\ttfamily\small\raggedright
You are an experienced emergency physician responsible for determining the most likely diagnosis for a patient presenting with abdominal symptoms.

You will be provided with the patient's clinical information, which may include the patient's history, symptoms, and results from previously completed diagnostic examinations. Consider all currently available clinical evidence before making the decision.

Based on the available evidence, identify the most likely diagnosis among the following conditions:

- appendicitis

- cholecystitis

- diverticulitis

- pancreatitis

In addition, assess whether the currently available information is sufficient to support a confident clinical judgment. If the evidence is adequate to support a reliable diagnosis, indicate that you are confident. Otherwise, indicate that additional diagnostic information may still be needed.

Example:

Patient history:
A 45-year-old male presents with severe abdominal pain and fever.

Laboratory results:
Elevated white blood cell count.

Imaging results:
CT scan shows an inflamed appendix.

Clinical assessment:
The combination of abdominal pain, fever, leukocytosis, and imaging evidence of an inflamed appendix strongly supports appendicitis, and the current findings are sufficient for diagnosis.

Answer:

Diagnosis: appendicitis

Confident: YES

Now evaluate the following patient case.

Patient description:\{current patient information\}

Return your answer in the following format:

Diagnosis: <appendicitis | cholecystitis | diverticulitis | pancreatitis>
Confident: <YES | NO>

\end{tcolorbox}
% \end{figure*}

\begin{table*}[htbp]
\centering
\small
\setlength{\tabcolsep}{4pt}
\caption{Performance comparison of our proposed LDTL with additional commercial models. (For proprietary LLMs, we perform evaluation through inference-only API calls without using the data for model training, fine-tuning, or retention. This protocol follows the PhysioNet data use agreement and ensures that no protected clinical data are incorporated into external model training.) }
\label{extraEx}
\begin{tabular}{lccccccccc}
\toprule
 & \multicolumn{5}{c}{Accuracy} 
 & \multirow{2}{*}{F1-Score} 
\\
\cmidrule(lr){2-6} 
Method 

& Appendicitis & Cholecystitis & Diverticulitis & Pancreatitis & \textbf{Overall} 
 \\
\midrule

SFT-all (llama-3-8b)
& 97.9 & 93.1 & 90.4 & 90.7 & 94.2 
& 95.8 &
\\

\midrule

gpt-4o \cite{openai2024gpt4o}
& 99.4 & 98.4 & 88.4 & 88.8 & 95.6 
& 95.7
\\

gpt-4o-mini \cite{openai2024gpt4omini}
& 94.7 & 91.5 & 82.6 & 86.1 & 90.5 
& 90.6 
\\

claude-sonnet-4 \cite{anthropic2025claude4}
& 99.3 & 98.4 & 90.3 & 88.8 & 95.8
& 95.8 
\\

gemini-3-flash \cite{gemini2024report}
& 99.4 & 94.6 & 90.3 & 90.7 & 95.2
& 95.3
\\

deepseek-chat \cite{deepseek2024llm}
& 99.4 & 97.6 & 92.3 & 92.5 & 96.6 
& 96.6 
\\

\midrule
LDTL
& 98.9 & 95.4 & 88.8 & 87.9 & 93.4 
& 91.7 
\\

\bottomrule
\end{tabular}

\end{table*}

\begin{multicols}{2}
\section{Dataset Preprocessing and Training Details }
\label{appendix:data}
We pre-process the dataset to preserve the benchmark's sequential semantics while standardizing inputs for LLM-based agents. First, when repeated measurements of the same clinical test were available within a single hospitalization, we retained only the earliest recorded value to approximate a realistic early stage diagnostic setting. Second, MIMIC-CDM provides standardized mappings of test names across patients, which enables consistent action definitions in our decision framework. This normalization is essential for modeling test selection, as inconsistencies in clinical documentation would otherwise prevent consistent querying of the same test across different cases. \citep{baniharouni2025lacdm} Third, we define the planning action space as a set of clinically meaningful test categories, including physical examination, imaging studies, and laboratory tests. When an action is selected, its associated results are appended to the current patient context in a structured textual format. To the best of our knowledge, MIMIC-CDM is currently the only publicly accessible dataset that supports simulation of hypothesis-driven sequential clinical decision making. We adopt the official patient-level split, using 70\% of cases for training, 10\% for validation, and 20\% for testing.

We adopt Llama-3 8B as the backbone language model and perform parameter-efficient fine-tuning using LoRA. 
The LoRA rank is set to 16 with a scaling factor of 32, and the target modules to $q\_proj$ and $v\_proj$, with a dropout rate of 0.1. We fine-tune the models using the AdamW optimizer with a learning rate of $2\times10^{-5}$. We use a per-device batch size of 1, with gradient accumulation steps of 8 for the diagnoser and 1 for the planner to balance memory efficiency and optimization stability. We set the temperature parameter $\tau=1.0$ for constructing the action preference distribution. And we choose 5 random seeds 17, 37, 42, 123, and 678. All experiments are conducted on NVIDIA H100 GPUs. Training typically requires approximately 6 hours on two H100 GPUs.

\begin{figure*}[htbp]
\centering
\includegraphics[width=0.97\textwidth]{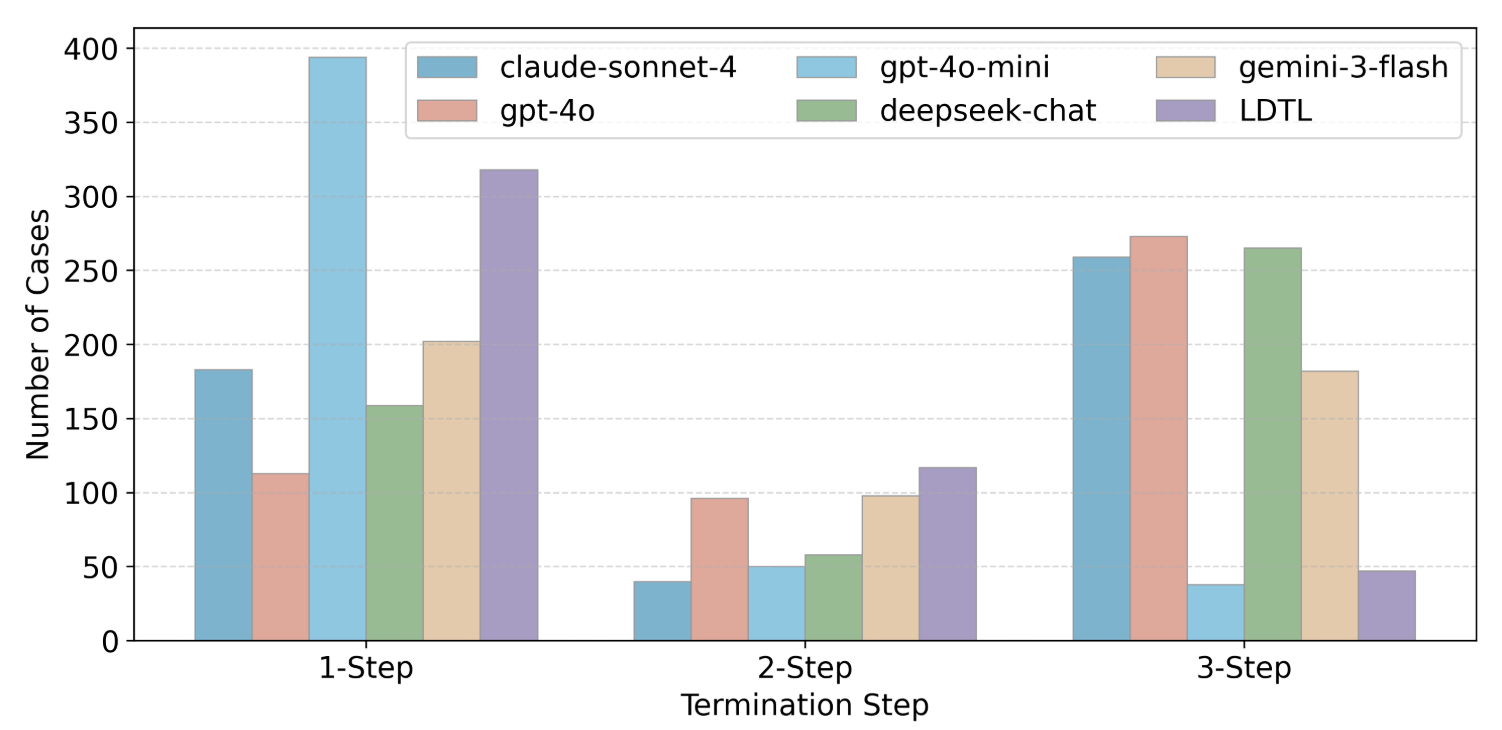}\vspace{-3mm}
\caption{Distribution of diagnostic termination steps on the test cases. Numbers indicate how many cases terminate after one, two, or three actions.}
\label{fig:ex_termination_steps}\vspace{-5mm}
\end{figure*}

\section{Additional Model Comparisons}
\label{appC}

We further compare our method with several widely used commercial LLMs, including GPT-4o, GPT-4o-mini, Claude-Sonnet-4, Gemini-3-Flash, and DeepSeek-Chat. These models are used purely for inference within the same sequential diagnostic framework. At each step, the model predicts the next diagnostic action and updates the disease prediction based on the currently available evidence. In this setting, the planning stage is implemented as direct next action prediction by the LLM, without latent path modeling or trajectory-level 
regularization. As shown in Table \ref{extraEx}, these models achieve strong predictive performance, with mean accuracy between 95.2 and 96.6 and macro F1 scores around 95.

Our method LDTL achieves a mean accuracy of 93.4 and a macro F1 score of 91.7. While slightly lower than the strongest proprietary models, it is important to note that these commercial systems typically rely on substantially larger model scales and extensive pretraining corpora. As a result, they may benefit from stronger general medical knowledge and prior exposure to similar clinical descriptions. In contrast, LDTL is built on a significantly smaller open source model llama-3-8b and trained for this sequential diagnostic task. Despite this difference in model capacity, our approach achieves comparable performance, highlighting the effectiveness of the proposed framework.

Figure \ref{fig:ex_termination_steps} compares the distribution of termination steps across different models. Our method LDTL frequently reaches a diagnosis in early stages, with a large portion of cases terminating after a single step with 318 cases, suggesting that the model is able to identify informative evidence early in the diagnostic process. At the same time, it continues to request additional examinations when uncertainty remains, resulting in a moderate number of two-step decisions.
In contrast, several strong LLM baselines such as GPT-4o, Claude-Sonnet-4, and DeepSeek-Chat tend to rely more heavily on three-step diagnostic sequences, indicating a more conservative strategy that gathers additional evidence before committing to a diagnosis. Gemini-3-Flash shows a more balanced distribution across one, two, and three-step terminations. Besides, GPT-4o-mini shows a strong bias toward one-step termination, suggesting a preference to make early decisions with minimal evidence.

Overall, these patterns highlight differences in diagnostic behavior across models, where LDTL demonstrates a more efficient evidence acquisition strategy while still allowing additional diagnostic steps when necessary.
\end{multicols}

\end{document}